\documentclass[journal]{IEEEtran}
\usepackage{amsmath,amsfonts}
\usepackage{algorithmic}
\usepackage{array}
\usepackage[caption=false,font=footnotesize,labelfont=sf,textfont=sf]{subfig}
\usepackage{textcomp}
\usepackage{stfloats}
\usepackage{url}
\usepackage{verbatim}
\usepackage{graphicx}
\usepackage{tabularx}
\usepackage{makecell}
\usepackage[font=normalsize]{caption}
\usepackage{tcolorbox}
\usepackage{layouts}
\usepackage{colortbl}
\usepackage{csquotes}
\usepackage{booktabs}
\usepackage{placeins}
\usepackage[]{todonotes}
\def\BibTeX{{\rm B\kern-.05em{\sc i\kern-.025em b}\kern-.08em
		T\kern-.1667em\lower.7ex\hbox{E}\kern-.125emX}}
\usepackage{balance}
\usepackage[noabbrev, capitalize]{cleveref}

\usepackage{siunitx}

\DeclareSIUnit \decibel {dB}
\usepackage{acro}
\DeclareAcronym{dl}{
	short=DL,
	long=deep learning,
}

\DeclareAcronym{dip}{
	short=DIP,
	long=Deep Image Prior,
}

\DeclareAcronym{deco-dip}{
	short=DECO-DIP,
	long=DECOnvolution Deep Image Prior,
}

\DeclareAcronym{psf}{
	short=PSF,
	long=point spread function,
}

\DeclareAcronym{rl}{
	short=RL,
	long=Richardson-Lucy,
}

\DeclareAcronym{tder}{
	short=TD ER, 
	long=time-dependent entropy deconvolution,
}

\DeclareAcronym{nlm}{
	short=NLM,	
	long=non-local means,
}

\DeclareAcronym{bm3d}{
	short=BM3D,
	long=block-matching 3D,
}

\DeclareAcronym{snr}{
	short=SNR,
	long=signal-to-noise ratio,
}

\DeclareAcronym{psnr}{
	short=PSNR,
	long=peak signal-to-noise ratio,
}

\DeclareAcronym{cnn}{
	short=CNN,
	long=convolutional neural network,
}

\DeclareAcronym{relu}{
	short=ReLU,
	long=rectified linear unit,
}

\DeclareAcronym{gan}{
	short=GAN,
	long=generative adversarial network,
}

\DeclareAcronym{sim}{
	short=SIM,
	long=structured illumination microscopy,
}

\DeclareAcronym{rdpv}{
	short=RDPV,
	long=Recursive Deep Image Prior Video,
}
\DeclareAcronym{ssim}{
	short=SSIM,
	long=structural similarity index measure,
}
\DeclareAcronym{lpips}{
	short=LPIPS,
	long=learned perceptual image patch similarity,
}

\DeclareAcronym{umap}{
	short=UMAP,
	long=Uniform Manifold Approximation and Projection
}

\DeclareAcronym{dbscan}{
	short=DBSCAN,
	long=Density-Based Spatial Clustering of Applications with Noise
}

\DeclareAcronym{ccd}{
	short=CCD,
	long=charge-coupled device,
}
\DeclareAcronym{pmt}{
	short=PMT,
	long=photomultiplier tube,
}
\DeclareAcronym{mos}{
	short=MOS,
	long=metal-oxide-semiconductor,
}

\DeclareAcronym{roi}{
	short=ROI,
	long=region of interest,
}

\DeclareAcronym{mse}{
	short=MSE,
	long=mean squared error,
}

\DeclareAcronym{mae}{
	short=MAE,
	long=mean absolute error,
}

\DeclareAcronym{gt}{
	short=gt,
	long=ground truth,
}

\DeclareAcronym{fmd}{
	short=FMD,
	long=Fluorescence Microscopy Denoising,
}

\DeclareAcronym{mito}{
	short=mito,
	long=mitochondria,
}

\DeclareAcronym{bpae}{
	short=BPAE,
	long=bovine pulmonary artery endothelial,
}

\DeclareAcronym{na}{
	short=NA,
	long=numerical aperture,
}

\DeclareAcronym{nas}{
	short=NAS,
	long-plural=es,
	short-plural=,
	long=neural architecture search,
}

\DeclareAcronym{ccp}{
	short=CCP,
	long-plural=s,
	short-plural=s,
	long=clathrin-coated pit,
}

\DeclareAcronym{er}{
	short=ER,
	long=endoplasmic reticulum,
}

\DeclareAcronym{mt}{
	short=MT,
	long=microtubules,
}

\DeclareAcronym{w2s}{
	short=W2S,
	long=Widefield2SIM,
}
\DeclareAcronym{fov}{
	short=FOV,
	long=fields of view,
}

\DeclareAcronym{tv}{
	short=TV,
	long=Total Variation,
}

\DeclareMathOperator*{\argmin}{arg\,min}

\begin{document}
\title{Automating Parameter Selection in Deep Image Prior for Fluorescence Microscopy Image Denoising via Similarity-Based Parameter Transfer}
\author{
    \IEEEauthorblockN{
    Lina Meyer\IEEEauthorrefmark{1}, 
    Felix Wissel\IEEEauthorrefmark{1}, 
    Tobias Knopp\IEEEauthorrefmark{2}, 
    Susanne Pfefferle\IEEEauthorrefmark{3},
    Ralf Fliegert\IEEEauthorrefmark{4},
    Maximilian Sandmann\IEEEauthorrefmark{4},
    Liana Uebler\IEEEauthorrefmark{4},
    Franziska Möckl\IEEEauthorrefmark{4},
    Björn-Philipp Diercks\IEEEauthorrefmark{4},
    David Lohr\IEEEauthorrefmark{1}, 
    René Werner\IEEEauthorrefmark{1}\\[0.3cm]}
    \IEEEauthorblockA{\IEEEauthorrefmark{1}Institute for Applied Medical Informatics, Institute of Computational Neuroscience, Center for Biomedical Artificial Intelligence (bAIome), University Medical Center Hamburg-Eppendorf, Germany\\}
    \IEEEauthorblockA{\IEEEauthorrefmark{2}Institute for Biomedical Imaging, Hamburg University of Technology, Hamburg, Germany; Department of Biomedical Imaging, University Medical Center Hamburg-Eppendorf, Germany; Fraunhofer Research Institution for Individualized Medical Technology and Engineering IMTE, Lübeck, Germany\\}
    \IEEEauthorblockA{\IEEEauthorrefmark{3}Institute of Medical Microbiology, Virology and Hygiene, University Medical Center Hamburg-Eppendorf, Germany\\}
    \IEEEauthorblockA{\IEEEauthorrefmark{4}Institute of Biochemistry and Molecular Cell Biology, University Medical Center Hamburg-Eppendorf, Germany\\[0.3cm]}
\thanks{This study was supported by the Deutsche Forschungsgemeinschaft (DFG), Project-ID: 335447727 – SFB 1328 (Projects A02 and A18). \textit{David Lohr and René Werner are shared last authors. Corresponding authors: David Lohr, René Werner.}\\
%
This article has supplementary material provided by the authors available.
}}

	

\maketitle

\begin{abstract}
  Unsupervised deep image prior (DIP) addresses shortcomings of training data requirements and limited generalization associated with supervised deep learning. The performance of DIP depends on the network architecture and the stopping point of its iterative process. Optimizing these parameters for a new image requires time, restricting DIP application in domains where many images need to be processed. Focusing on fluorescence microscopy data, we hypothesize that similar images share comparable optimal parameter configurations for DIP-based denoising, potentially enabling optimization-free DIP for fluorescence microscopy. 
  We generated a calibration (n=110) and validation set (n=55) of semantically different images from an open-source dataset for a network architecture search targeted towards ideal U-net architectures and stopping points. The calibration set represented our transfer basis. The validation set enabled the assessment of which image similarity criterion yields the best results. We then implemented AUTO-DIP, a pipeline for automatic parameter transfer, and compared it to the originally published DIP configuration (baseline) and a state-of-the-art image-specific variational denoising approach. 
  We show that a parameter transfer from the calibration dataset to a test image based on only image metadata similarity (e.g., microscope type, imaged specimen) leads to similar and better performance than a transfer based on quantitative image similarity measures. 
  AUTO-DIP outperforms the baseline DIP (DIP with original DIP parameters) as well as the variational denoising approaches for several open-source test datasets of varying complexity, particularly for very noisy inputs. Applications to locally acquired fluorescence microscopy images further proved superiority of AUTO-DIP. Thus, AUTO-DIPs dictionary-like function improves DIP-based denoising speed and quality, enabling routine applications in the domain of fluorescence microscopy imaging.

	\end{abstract}
	

	\section{Introduction}
	
	\noindent Fluorescence microscopy is widely used to study biological processes~\cite{fluorescence-microscopy}. However, the resulting images often exhibit a low \ac{snr} due to short exposure times, which are required to limit photodamage and to capture rapid dynamics at high frame rates. This makes post-acquisition denoising indispensible~\cite{noise}. 
	
	State-of-the-art denoising algorithms are predominantly based on supervised \ac{dl}~\cite{Weigert_Schmidt_Boothe_Mueller_Dibrov_Jain_Wilhelm_Schmidt_Broaddus_Culley_et_al, Chen_Sasaki_Lai_Su_Liu_Wu_Zhovmer_Combs_Rey_Suarez_Chang_et_al_2021}. A key limitation is the need for large datasets containing pairs of high- and low-\ac{snr} images to train them. Moreover, since completely noise-free images cannot be acquired, the ground truth images for training and evaluation are typically approximated, for instance, using long exposure times or by averaging multiple noisy images. Furthermore, models trained on a specific dataset often fail to generalize to other types of microscopy images~\cite{noise}.
	
	Classical denoising methods avoid these issues by relying on explicit image priors to regularize the inverse problem. They formulate denoising as an optimization task, balancing a data fidelity term with a regularization term that encodes assumptions about the image structure. Common examples include Tikhonov regularization~\cite{tikhonov1977solutions}, which penalizes large variations in pixel intensities, and \ac{tv} regularization~\cite{Rudin_Osher_Fatemi_1992}, which promotes piece-wise smoothness. More advanced methods, such as \ac{bm3d}~\cite{bm3d}, exploit self-similarity by grouping similar patches of an image and filtering them jointly. While effective in many cases, these methods strongly depend on the chosen regularizer and often struggle to capture the complexity of natural images.
	
	
	To circumvent the need for large training datasets and handcrafted priors, we use the unsupervised Deep Image Prior (DIP)~\cite{dip} for denoising. Instead of relying on an explicit prior, DIP employs the architecture of a randomly initialized convolutional neural network as an implicit regularizer. The network weights are optimized to reconstruct the corrupted (i.e, the low-SNR) image. Due to its structural bias, the network tends to capture natural image structures before overfitting to noise, enabling denoising by stopping the training process early.
	
	%
	%
    As shown by Ulyanov et al.~\cite{dip}, the denoising performance depends on the network architecture and the stopping point. Finding the optimal parameters for each image requires an exhaustive search, which is computationally expensive and time-consuming. 
	Existing approaches address network architecture searches to identify general or image-specific optimal architectures~\cite{Chen_Gao_Robb_Huang_2020, Arican_Kara_Bredell_Konukoglu_2022}, optimal stopping point detection~\cite{Li_Zhuang_Liang_Peng_Wang_Sun_2021, Jo_Chun_Choi_2021, Wang_Li_Zhuang_Chen_Liang_Sun_2021}, and combinations thereof~\cite{Ho_Gilbert_Jin_Collomosse_2021}; for details see Section~\ref{sec:methods}. However, most of them target natural images instead of fluorescence microscopy images and therefore found solutions for very different dynamic intensity ranges as well as different noise and sparsity characteristics and overall image complexity. In addition, none combines network architecture and stopping point determination while providing a fast and easy way to select parameters for unseen images. 
	
	Focusing on fluorescence microscopy, we hypothesize that similar images share comparable optimal parameter configurations for DIP-based denoising, and we propose AUTO-DIP, a strategy that determines both network architecture and stopping point by transferring parameters based on image similarity.
	To test our hypothesis, we evaluate whether transferring parameters from similar images yields denoising performance comparable to that achieved through image-specific (i.e., for the individual image) parameter optimization. We systematically compare different similarity measures -- general properties such as microscope and specimen type; different pixel-based and semantic similarity measures, and a combination thereof -- to identify the most effective criteria for guiding parameter transfer. 
    We demonstrate that similarity-based architecture and stopping point selection improves DIP-based denoising of fluorescence microscopy images compared to static parameter selection. In addition, applying AUTO-DIP results in 3-times faster denoising than using the original DIP parameters. 
    A comparison to a state-of-the-art unsupervised variational denoising approach further illustrates the denoising potential of AUTO-DIP, which deals well with varying image complexity. 

	\section{Methods\label{sec:methods}}
	
	\subsection{Deep Image Prior and Related Work}
	
	\noindent Image restoration is commonly formulated as an inverse problem solved by minimizing an energy function combining a data fidelity and a regularization term:
	\begin{align}
		\label{0000-eq:energymin}
		x^*=\argmin_x E\left(x,x_o\right)+R\left(x\right),
	\end{align}
	\noindent where \(x_0\) is the degraded input image, \(x\) the restored image, \(E\) a task-specific data fidelity term, and \(R\) an explicit regularizer~\cite{dip}. 
 
    Unlike classical methods, DIP replaces the explicit regularizer \(R\) with an implicit prior encoded by a \ac{cnn} \(f_{\theta}\) initialized with random weights \(\theta\). The restoration is performed by optimizing the network parameters:
	
	\begin{align}
		\label{0000-eq:networkparam}
		\theta^*=\argmin_\theta E\left(f_\theta\left(z\right),x_o\right)
	\end{align}
	
	\noindent where \(z\) is a fixed random input tensor. For denoising, \(E\) is typically the squared \(L^2\) norm between network output and corrupted image.
	
	The central hypothesis of DIP is that the CNN architecture imposes a structural bias that favors restoring natural image features over noise. Consequently, during training, the network first learns to represent the main image structures before fitting noise. Early stopping of the training prevents overfitting, yielding a denoised output \(x^*=f_{\theta}(z)\).
	
	The standard architecture used in DIP is a U-net~\cite{dip}, originally introduced for biomedical image segmentation~\cite{unet}. U-nets consist of a symmetric encoder-decoder structure: the encoder (downsampling path) reduces spatial resolution through strided convolutions, progressively extracting higher-level features, while the decoder (upsampling path) reconstructs the image by restoring resolution. The number of downsampling and upsampling operations defines the depth of the network, while the number of feature maps used to represent the image at different layers can be interpreted as the width of the network. 
	Further, a key architectural feature of U-nets is the use of skip connections, which link corresponding layers in the encoder and decoder paths. These connections allow low-level spatial details lost during downsampling to be directly reintroduced during reconstruction, improving the preservation of fine image structures.
	
	As outlined in the introduction, several studies have already investigated how to optimize either the network architecture or the stopping point for DIP. 
	From a frequency perspective, Chakrabarty and Maji showed that convolutional networks naturally prioritize low-frequency components, which explains why early stopping acts as a form of low-pass filtering~\cite{spectral_bias}. They also found that deep and narrow networks decouple frequencies more effectively than wide and shallow ones. Liu et al. further recommended architectural configurations based on image texture: fine-grained images benefit from shallow, wide networks, while coarse-grained ones perform better with deep, narrow architectures. They also noted that skip connections resemble the effect of shallower networks, except in very deep models~\cite{Liu_Li_Pang_Nie_Yap_2023}.
	
	Other work used \ac{nas} to automate the architecture design. NAS-DIP~\cite{Chen_Gao_Robb_Huang_2020} applies reinforcement learning to optimize upsampling methods, kernel sizes, and skip connections for a given dataset. Ho et al. instead performed image-specific search using a genetic algorithm~\cite{Ho_Gilbert_Jin_Collomosse_2021}. Their findings suggested that optimal architectures cluster by the visual style of the considered images (e.g., vector art, oil painting, or pen and ink images). ISNAS-DIP~\cite{Arican_Kara_Bredell_Konukoglu_2022} proposes an optimal DIP architecture by evaluating the spectral similarity between the input image and the network output after one DIP iteration for randomly sampled network architectures. 
 
	
	Proposed approaches to target the optimal stopping point in DIP training include learning-based predictors~\cite{Li_Zhuang_Liang_Peng_Wang_Sun_2021}, overfitting detection via effective degrees of freedom~\cite{Jo_Chun_Choi_2021}, and stopping based on output variance~\cite{Wang_Li_Zhuang_Chen_Liang_Sun_2021}. Further, regularization terms have been added to the loss to suppress overfitting, including explicit penalties like total variation~\cite{Mataev_Milanfar_Elad,Cascarano_Franchini_Kobler_Porta_Sebastiani_2023}, or implicit ones using classical denoising approaches~\cite{Sun_Latorre_Sanchez_Cevher_2021}.

	%

	\subsection{Our Approach: AUTO-DIP}\label{approach}
	
	\noindent The mentioned works were based on common natural image data sets; in contrast, our work focuses on a specific image type: fluorescence microscopy images. We hypothesize that it is possible to transfer optimal DIP parameters between similar fluorescence microscopy images. While existing approaches focused either on optimal architecture selection or stopping point determination, we jointly optimized architectural parameters -- namely depth, width, and the presence of skip connections of a U-net architecture -- as well as the number of training iterations for dedicated calibration and validation sets, which contained a selection of fluorescence microscopy images from a publicly available dataset dedicated to denoising of microscopy images: the FMD dataset \cite{Zhang_Zhu_Nichols_Wang_Zhang_Smith_Howard_2019}. For dataset details see \ref{data}.
 
    In short, the idea of AUTO-DIP is to compare new input images to fluorescence microscopy images from the calibration set using one or more similarity measures. It then selects the most similar calibration image and applies DIP-based denoising using the parameter configuration that was found to be optimal for that calibration set image during the grid search. Experiments to identify which similarity measures provide the best denoising results are detailed later.
 
    The search space was defined as: 
	\begin{itemize}
		\item network depth: \(d \in \{4,5,6,7,8\}\)
		\item network width: \(w \in \{16,32,64,128,256,512\}\)\\ 
        Here, $w$ means that the same number $w$ of feature maps was used at each network level.\\ 
        Furthermore, we integrated a network architecture with progressively increasing width across layers, reaching $512$ feature maps at the deepest layer, into the optimization. This configuration is referred to as \(v_{512}\). For a network of depth $d$, the number of feature maps at depth level $d$ is computed as \(w_j = 2^{(9-d+j)}\)
		\item skip connections: yes or no
		\item iterations until stopping: \(i \in \{100, 200, ..., \num{3000}\}\)
	\end{itemize}
    For the transfer of parameter configurations derived by this grid search, we evaluated the following image similarity criteria on both the calibration and validation sets:
	\begin{enumerate}
		\item \emph{Group-Based Similarity}: Parameter transfer based on image metadata similarity, that is, parameters were transferred based on the microscope type, the same specimen type, and both the same microscope and specimen type. 
        Here, optimal parameters refer to those that, on average, yield the best denoising performance for the specific groups of the calibration dataset.  
		\item \emph{Metric-Based Similarity}: Transfer of the optimal parameters from the most similar image in the calibration set. We evaluated the following pixel-based and perceptual similarity measures: \ac{mae}, \ac{mse}, \ac{psnr}, \ac{ssim}, \ac{lpips} \cite{Zhang_Isola_Efros_Shechtman_Wang_2018},  and Euclidean distance in a UMAP representation computed based on the calibration images.
		\item \emph{Combined Group-Metric Similarity}: Combination of the group-based and the metric-based approach; the pool of the calibration images that is considered for metric-based image-to-image similarity comparison is restricted to images from the same group.
	\end{enumerate}

    \subsection{Calibration and Validation Dataset Definition}
    Image selection for the calibration and validation dataset was based on three criteria: microscope type, specimen, and semantic similarity. The later was approximated using Euclidean distances between UMAP embeddings of the individual images (see Section S2.1). These distances were also assessed as a similarity measure to select parameter configurations for AUTO-DIP (section \ref{approach}).      
    
    
	\subsection{Microscopy Datasets}\label{data}   
    To test generalizability and evaluate the limitations of our approach, we compiled images from the FMD and five other datasets for testing. The datasets are summarized below; further details can be found in the respective publications.
	
	\subsubsection{FMD}\label{fmd}
	The \acf{fmd} dataset is a collection of \num{12000} fluorescence microscopy images, acquired with three microscope types: confocal, widefield, and two-photon. The dataset includes three specimen types: zebrafish, mouse brain tissue, and \ac{bpae} cells. For \ac{bpae} cells, the components actin, mitochondria, and nucleus were imaged individually using three different wavelengths. 
	A total of 12 specimens with 20 fields of view were captured 50 times. Different noise levels were simulated by averaging a varying number $S$ of images (\(S = 1, 2, 4, 8, 16\)). The 50 images were averaged to produce a high-quality reference image, serving as the ground truth for subsequent analyses.  Each image is an 8-bit grayscale image with a resolution of \(512 \times 512\) pixels. 
 
    For the calibration dataset, 110 images were selected, and 55 images were included into the validation set (see Tables S1 to S6). The noisy images of the dataset (that is, the model inputs) corresponded to the highest noise levels (\(S=1,2\)), and the \ac{bpae} cell parts (actin, mitochondria, nucleus) were used to include varying sample complexity.     
	
	For model evaluation, the designated FMD mixed test set~\cite{Zhang_Zhu_Nichols_Wang_Zhang_Smith_Howard_2019} was used as a test set. It contains all specimens and noise levels, totaling 240 images.
	
	
	\subsubsection{Hagen}
	The denoising dataset provided by Hagen et al. \cite{Hagen_Bendesky_Machado_Nguyen_Kumar_Ventura_2021} contains 567 pairs of high-SNR and low-SNR fluorescence microscopy images. The noise levels were varied by adjusting exposure time, laser power, and detector gain during data acquisition. The dataset includes images of actin, mitochondria, and nuclei in \ac{bpae} cells, acquired using widefield and confocal microscopes. Image sizes range from \(512 \times 512\) to \(2048 \times 2048\) pixels, all with 16-bit image intensity range.
	We used the dataset solely as a test set, employing the dedicated 96-image test subset defined by Hagen et al. \cite{Hagen_Bendesky_Machado_Nguyen_Kumar_Ventura_2021}. 
	
	\subsubsection{BioSR}
	The super-resolution microscopy dataset  BioSR~\cite{Qiao2020} by Qiao et al. consists of over \num{2200} pairs of low-resolution (widefield microscopy) and high-resolution (corresponding structured illumination microscopy) images. It  contains four structures: \acp{ccp}, \ac{er}, actin and microtubules, each imaged with nine different signal levels. 
	Image size is \(512 \times 512\) pixels and the dynamic range is 16-bit.
	We used one widefield microscopy image per structure, each with five different noise levels (see Table S7) and the highest available signal level as a ground truth image for testing purposes.
	
	\subsubsection{W2S}
	\ac{w2s} is a joint denoising and super-resolution dataset containing 360 (120 \ac{fov} for three different biological structures) 
    image sets of human cell components~\cite{zhou2020w2smicroscopydatajoint}. Each set includes widefield low-resolution noisy images of five noise levels, a low-resolution noise-free image, and a high-resolution \ac{sim} image. The five noise levels were generated by averaging 1, 2, 4, 8, or 16 noisy images, and a noise-free version of the images is approximated by averaging 400 images of the same \ac{fov}. 
	All images have a size of \(512 \times 512\) pixels with a 16-bit intensity range. We only used the low-resolution images of one distinct \ac{fov} (all three biological structures) with the five different noise levels (see Table S9).
	
	\subsubsection{Shah}	
	The dataset for joint denoising and super-resolution provided by Shah et al. contains 101 \acp{fov} of human tubulin filaments~\cite{Shah_2024}. Each \ac{fov} was captured at 200 time points for 5 orientations and 3 phases using widefield microscopy. The \ac{snr} decreases with every timestamp due to photobleaching. All images are of size \(512 \times 512\) pixels with a 16-bit dynamic range. Different noise levels were generated by averaging 1, 2, 3, 4, and 6 images acquired at a specific timestamp, while the ground truth was generated by averaging 9 images. For testing purposes, two \acp{fov} with these five different noise levels were used (see Table S8). 
	

	\subsubsection{UKE-internal Data}
     In addition to the above public datasets, we evaluated the AUTO-DIP performance using internally acquired fluorescence microscopy image sets. For the first sample, A549-A/T cells~\cite{meister_mycophenolic_2025}
     were grown to 80\% confluency on coverslips and infected with recombinant SARS-CoV-2 as described previously~\cite{fliegert_targeting_2025,lutgehetmann_generation_2025}. Details regarding sample preparation, staining of nuclei, and antibodies are provided in the supplemental material and Tables S10-11. For imaging, a super-resolution spinning-disk microscope (Visitron) equipped with a CSU-W1 SoRa Optic (2.8x, Yokogawa) was used. Laser settings are described in Table S12. Images were captured using a sCMOS camera (Orca-Flash 4.0, C13440-20CU Hamamatsu) and acquired with VisiView software by Visitron. 
		
	\subsection{Data Preprocessing}
	
	\noindent For multi-channel images, each channel was processed independently. Before denoising, each image was rescaled to a [0,1] value range based on its minimum and maximum intensity values. Images larger than \(512 \times 512\) pixels were divided into patches of \(512 \times 512\) pixels, with 128 pixels overlap. After denoising, the entire images were reassembled based on the denoised patches,  averaging the values for overlapping image regions. 
	
	In the Hagen et al. and BioSR datasets, noisy and ground truth images exhibit slight spatial shifts and differing dynamic ranges due to variations in microscope settings during acquisition. We therefore registered and adjusted the images before performance evaluation as described in the respective original dataset publications (see details in Section S2-B).
	
	
	\section{Experiments}
	\subsection{Optimal Parameter and Evaluation Measure Selection}
	\noindent For each image of the FMD calibration and validation datasets, we applied DIP for denoising using all parameter combinations of the defined search space. 
	We evaluated the image-specific performance for each parameter configuration based on \ac{mse}, \ac{mae}, \ac{psnr}, \ac{ssim}, and \ac{lpips}; the measures were evaluated for comparison of the denoised and the corresponding ground truth image. Thus, for each measure and image, the best parameter configuration was identified. 

    For the group-based identification of optimal DIP parameters, 15 subsets of the calibration image dataset were formed (3 sets according to the microscope type, 3 sets for the different specimens, 3$\times$3 combinations thereof). Optimal parameter sets for each group and each evaluation measure were determined by maximizing the average performance across all images of the group.

	As it was unclear during the conceptual phase which quantitative image similarity measure was best suited for evaluating denoising performance, we also investigated this aspect in detail. As a first step, we visually assessed the overall image impression of DIP output images from the calibration set that were 'optimal' according to the different measures. In a second step, the image complexity of the denoised images was quantitatively assessed by computing the mean image gradient and its difference to the mean image gradient of the corresponding ground truth image. A difference of zero would imply that the denoised image retains the structural complexity of the ground truth image.  
 
	\subsection{Parameter Transfer Evaluation: Validation Set}\label{subsec:validate}
	\noindent For each image of the FMD validation set, we comprehensively evaluated the different transfer strategies described in Sec.~\ref{approach}. 
	%
    The AUTO-DIP denoising performance for an individual image of the validation set was compared to
    \begin{enumerate}
        \item the best possible performance for the considered image, evaluated based on the grid search results for the specific image, and
        \item the performance using the parameters proposed in the original DIP publication by Ulyanov et al. (\(d=5\), \(w=128\), \(s=4\), \(i=1800\)), subsequently referred to as just 'DIP'. 
    \end{enumerate}

    The first approach determines the upper boundary of DIP-based denoising performance. The second defines a baseline performance, with the hypothesis that AUTO-DIP leads to a better performance than the original DIP parameters. 

    To further assess the performance of the (AUTO-)DIP approaches, we applied a state-of-the-art variational denoising method as an additional reference. We used a sparsity-based denoising~\cite{Zhao_Zhao_Li_Huang_Xing_Zhang_Qiu_Han_Shang_Sun_et_al._2022} specifically tailored to fluorescence microscopy images that  
    integrates second-order derivative regularization with sparsity constraints to suppress noise, while preserving structural details in fluorescence microscopy images. 
    For a fair comparison, we also performed a grid search to identify an optimal parameter configuration. 
    
	
	\subsection{Parameter Transfer: Test Sets}
	\noindent To evaluate the generalizability of the findings for the FMD validation set, we tested the best-performing transfer strategy on the five additional microscopy datasets described in \cref{data}. Similar to the validation set experiments, we applied 'DIP' configurations by Ulyanov et al., and (2) the best AUTO-DIP configurations to each image. Again, we also applied sparsity-based denoising as a reference. 
    We compared the denoising performance for the different approaches and investigated how image characteristics like noise level and sparsity affect our approach.

	\section{Results}\label{results}

    \noindent The source code for AUTO-DIP is available at \url{https://github.com/IPMI-ICNS-UKE/AUTO-DIP}. All experiments were conducted on a computer with an AMD EPYC 7513 32-core processor (3.6 GHz) and an NVIDIA A40 GPU (48 GB VRAM). The full parameter grid search for the FMD calibration and validation datasets required 33 days of wall-clock time. 
    Run times for the different DIP variants and test sets are provided in Table S14.
    On average, the transferred DIP configurations ran approximately three times faster than the original DIP parameter configuration. 

    \subsection{Optimal Parameter Identification and Evaluation Measure Selection on the Calibration Dataset}
    
     For the calibration set, the best DIP configurations according to MAE and SSIM resulted in very smooth outputs and the loss of almost all structural details within the cells, (see supplemental figures S2 and S3). DIP outputs for the best configurations in terms of MSE and PSNR still appeared very smooth. The best configurations in terms of LPIPS led to outputs that retained more details but also more noise. In addition, LPIPS does not guarantee pixel-wise intensity value agreement of denoised and reference images. A mixture of PSNR and LPIPS provided the best combination of high agreement in image complexity and visual perception such as structural detail as well as  pixel-level intensities between DIP outputs and the ground truth. 
     
     For identification of the best DIP configurations using both measures, we defined a ranking system. For each calibration set image, we ranked the quality of the DIP outcome for the different configurations with respect to both LPIPS (rank 1: lowest value) and PSNR (rank 1: highest value). The best configuration for a specific image was then determined by the lowest sum of both ranks. This combination led to smaller differences in the mean gradients of noise-reduced and reference images than all individual pixel-wise measures. 
    
\begin{figure*}
		\centering
		\includegraphics[width=\linewidth]{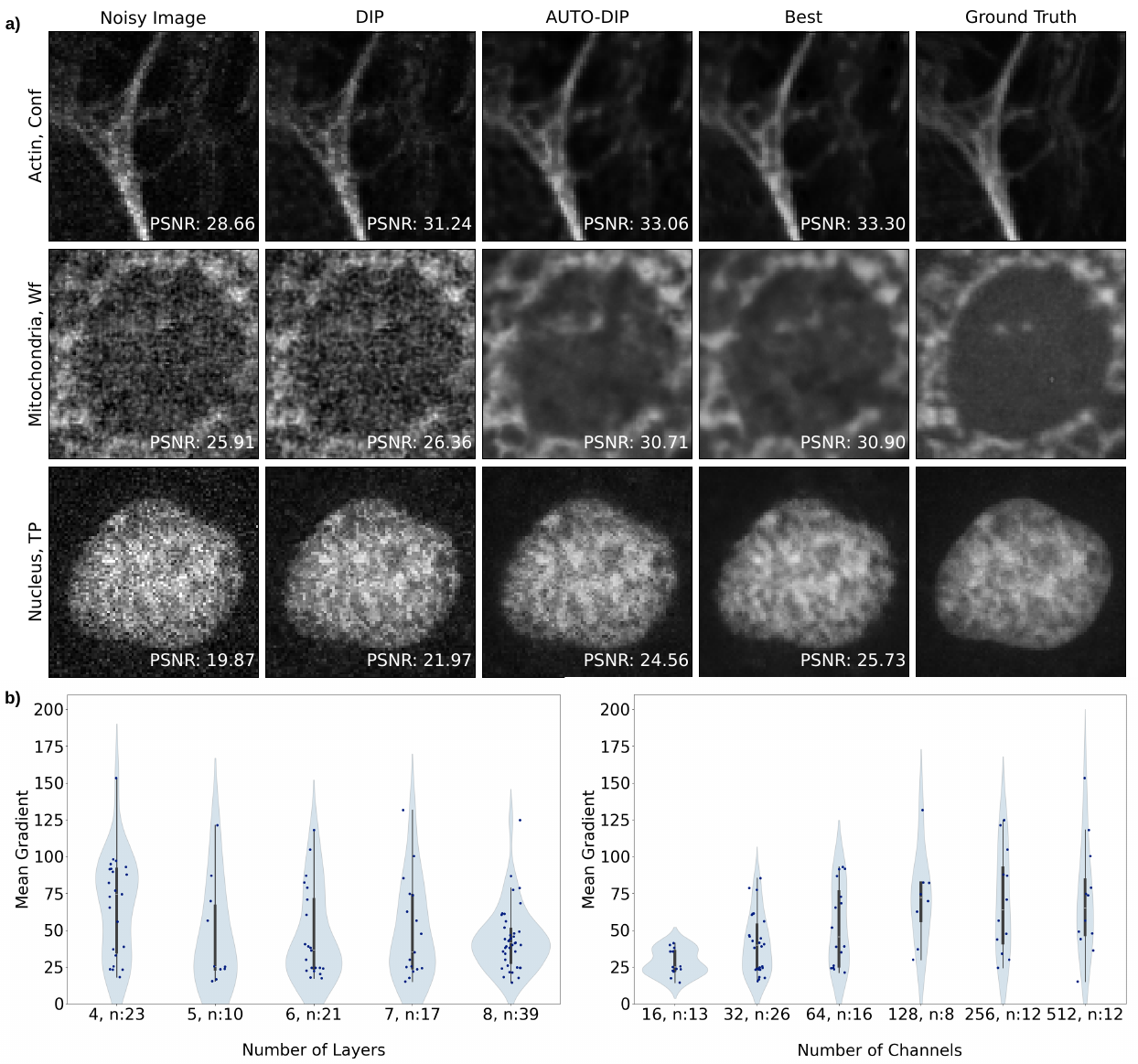}
		\caption{(a) Example results for the validation dataset. From left to right: noisy input image, result with original DIP configuration, AUTO-DIP result with transferred parameters using Group-Based Similarity, best result achievable from all parameters of the calibration set according to the combined LPIPS-PSNR measurement, ground truth image. Each row shows a \ac{roi} from a different image, covering the three structures and three microscope types contained of the training data.
        (b) Relationship of image complexity, measured by the mean image gradient, and optimal number of layers (left) and number of channels (right) for DIP-based denoising for the calibration set. The number n denotes the number of images for which the corresponding depth and width are found optimal in terms of the combined measurement of PSNR and LPIPS. \\
        Abbreviations: Conf -- Confocal, Wf -- Widefield, TP -- Two-Photon}
		\label{fig:train-results}
	\end{figure*}

    \cref{fig:train-results}a shows DIP-based  denoising results for selected calibration images, illustrating the improvement achieved by using per-image optimal configurations compared to the original DIP baseline. 
    Quantitatively, for the calibration dataset, the original DIP parameter configuration \cite{dip} led to a mean PSNR of 32.10 (higher = better) and a mean LPIPS of 0.205 (lower = better; zero is optimal). Selecting the optimal configuration as described yielded a mean PSNR of 35.56 and a mean LPIPS of 0.083.
    
    \cref{fig:train-results}b further illustrates the dependence of optimal network depth and height and the input image complexity, quantified by the mean image gradient. It shows a general trend: Images with higher mean gradients tend to benefit from shallower but wider networks, whereas lower-gradient images perform best with deeper network configurations but with a smaller number of feature maps per depth level. However, for network depth larger than 5 and network width above 128, the trend becomes only marginally apparent. 
    
    In addition, across almost all configurations, architectures with skip connections performed better. In contrast, the optimal number of training iterations varied widely by modality: the optimization runs for the  widefield images often converged within 300–500 iterations, while confocal and two-photon images required longer optimization, in some cases up to 2,900 iterations.
    	
	\subsection{Evaluate Parameter Transfer Strategies on Validation Set}

    After defining optimal DIP configurations for the calibration set images, we evaluated different image similarity criteria for parameter transfer between the calibration and validation dataset images. 
    \cref{tab:transfer_strats_comp} summarizes the results. Each cell shows the mean PSNR and mean LPIPS across all validation images. 

		\begin{table}
		\captionof{table}{Denoising performance for different parameter transfer strategies, measured by mean PSNR and LPIPS for the validation dataset images. The first column specifies the similarity measure used to find the `closest' image of the calibration dataset that is used for parameter selection. The remaining columns indicate the constraints applied to the nearest image search: \enquote*{Microscope-Specimen-Group} includes only images of the same specimen captured with the same microscope type, \enquote*{Microscope} includes images captured with the same microscope type, \enquote*{Specimen} includes images of the same specimen, and \enquote*{All} includes all images. Each cell contains the mean PSNR (top) and mean LPIPS values (bottom) for the FMD validation dataset images.
        '(Group Only)' refers to transfer of the, on average, best DIP configuration for the calibration images of the specific group; no quantitative similarity was evaluated between validation and calibration images. 
        Background color: lighter = better performance in terms of the average PSNR and LPIPS ranks across all entries of the table.  
		}
		\renewcommand{\arraystretch}{1.2}
		\begin{tabularx}{\linewidth}{p{1.4cm} | *{4}{>{\raggedleft\arraybackslash}X}}
			\toprule
			\raggedright\arraybackslash Similarity Measure &		\raggedright\arraybackslash{Microscope-Specimen Group} &		\raggedright\arraybackslash Microscope &		
			\raggedright\arraybackslash Specimen &
			\raggedright\arraybackslash All\\ 
			\midrule
			\makecell[l]{-- (Group \\  Only)}  & \cellcolor[rgb]{0.99,0.96,0.93} \hspace*{-0.35cm}
			\setlength\arrayrulewidth{1.2pt}
			\begin{tabular}{|p{1.26cm}|}
				\hline
				\makecell[r]{34.99 \\ 0.098 }\\
				\hline
			\end{tabular} & \cellcolor[rgb]{0.98,0.85,0.78} \makecell[r]{34.88 \\ 0.107} & \cellcolor[rgb]{0.91,0.57,0.65} \makecell[r]{34.17 \\ 0.127} & \cellcolor[rgb]{0.51,0.51,0.55} \makecell[r]{33.72 \\ 0.158} \\
			MAE & \cellcolor[rgb]{0.98,0.82,0.74} \makecell[r]{34.66 \\ 0.104} & \cellcolor[rgb]{0.98,0.72,0.65} \makecell[r]{34.56 \\ 0.114} & \cellcolor[rgb]{0.98,0.83,0.75} \makecell[r]{34.70 \\ 0.104} & \cellcolor[rgb]{0.96,0.63,0.62} \makecell[r]{34.40 \\ 0.119} \\
			MSE & \cellcolor[rgb]{0.98,0.82,0.74} \makecell[r]{34.67 \\ 0.104} & \cellcolor[rgb]{0.98,0.83,0.75} \makecell[r]{34.72 \\ 0.105} & \cellcolor[rgb]{0.98,0.83,0.75} \makecell[r]{34.70 \\ 0.103} & \cellcolor[rgb]{0.98,0.72,0.65} \makecell[r]{34.53 \\ 0.112} \\
			PSNR & \cellcolor[rgb]{0.98,0.82,0.74} \makecell[r]{34.67 \\ 0.104} & \cellcolor[rgb]{0.98,0.83,0.75} \makecell[r]{34.72 \\ 0.105} & \cellcolor[rgb]{0.98,0.83,0.75} \makecell[r]{34.70 \\ 0.103} & \cellcolor[rgb]{0.98,0.72,0.65} \makecell[r]{34.53 \\ 0.112} \\
			SSIM & \cellcolor[rgb]{0.98,0.80,0.72} \makecell[r]{34.67 \\ 0.105} & \cellcolor[rgb]{0.98,0.75,0.68} \makecell[r]{34.65 \\ 0.112} & \cellcolor[rgb]{0.94,0.59,0.63} \makecell[r]{34.32 \\ 0.119} & \cellcolor[rgb]{0.90,0.55,0.66} \makecell[r]{34.30 \\ 0.126} \\
			LPIPS & \cellcolor[rgb]{0.99,0.93,0.89} \makecell[r]{34.97 \\ 0.098} & \cellcolor[rgb]{0.99,0.92,0.88} \makecell[r]{34.97 \\ 0.100} & \cellcolor[rgb]{0.98,0.71,0.65} \makecell[r]{34.66 \\ 0.115} & \cellcolor[rgb]{0.97,0.71,0.64} \makecell[r]{34.65 \\ 0.115} \\
			\makecell[l]{Mean \\ Gradient} & \cellcolor[rgb]{0.99,0.95,0.92} \makecell[r]{34.98 \\ 0.098} & \cellcolor[rgb]{0.98,0.80,0.72} \makecell[r]{34.79 \\ 0.112} & \cellcolor[rgb]{0.98,0.88,0.82} \makecell[r]{34.93 \\ 0.107} & \cellcolor[rgb]{0.75,0.56,0.68} \makecell[r]{34.30 \\ 0.151} \\
			UMAP & \cellcolor[rgb]{0.99,0.91,0.86} \makecell[r]{34.88 \\ 0.101} & \cellcolor[rgb]{0.98,0.81,0.73} \makecell[r]{34.78 \\ 0.110} & \cellcolor[rgb]{0.98,0.87,0.80} \makecell[r]{34.79 \\ 0.103} & \cellcolor[rgb]{0.98,0.79,0.71} \makecell[r]{34.74 \\ 0.111} \\
			
			\bottomrule
		\end{tabularx}
		\label{tab:transfer_strats_comp}
	\end{table}

    The first row indicates the different groups considered for the group-based parameter strategies: microscope-specimen group (images of the same specimen, acquired with the same microscope type), microscope-only group, and specimen-only group. 
    The last column shows the quantitative results without consideration of the corresponding image metainformation. Thus, the cell 'Group Only' and 'All' refers to the transfer of the parameter configuration that led, on average, led to the best results across all calibration images, regardless of grouping. The following rows indicate the results for the image similarity-based strategies to select appropriate calibration images for parameter selection and transfer. For column 'All', the calibration image set for evaluation of the similarity measures spans the entire calibration set. For the other columns and groups, respectively, only the corresponding subset is used.

	Among all parameter transfer strategies, the group-based transfer for the most restricted group -- consideration of only images acquired with the same microscope type and imaging a similar specimen -- achieved the best performance, with a mean PSNR of 34.99. For comparison, the upper PSNR boundary, that is, the mean PSNR obtained for the image-specific optimal configuration after parameter grid search for the validation set, was 35.56. The lowest-performing approach was the naïve selection of the globally best average configuration for the calibration set, resulting in a mean PSNR of 33.72. Nonetheless, this still outperformed denoising performance using the original DIP configuration (31.90).
	
	Similar to the 'group only' transfer approach, the results for the metric-based parameter transfer indicate that restricting parameter transfer to images from the same microscope type, specimen type, or both improves denoising quality. However, a superiority of quantitative image similarity measures, that is, the selection of the optimal parameters of the most similar image of the calibration set, for automated parameter transfer cannot be shown. Therefore, we selected the microscope-specimen group transfer strategy for subsequent experiments. 
    The optimal configurations for each group according to the combined LPIPS-PSNR metric are listed in Table S13.
	
	\subsection{Parameter Transfer to Test Sets}

\begin{figure*}
		\centering
		\includegraphics[width=0.75\linewidth]{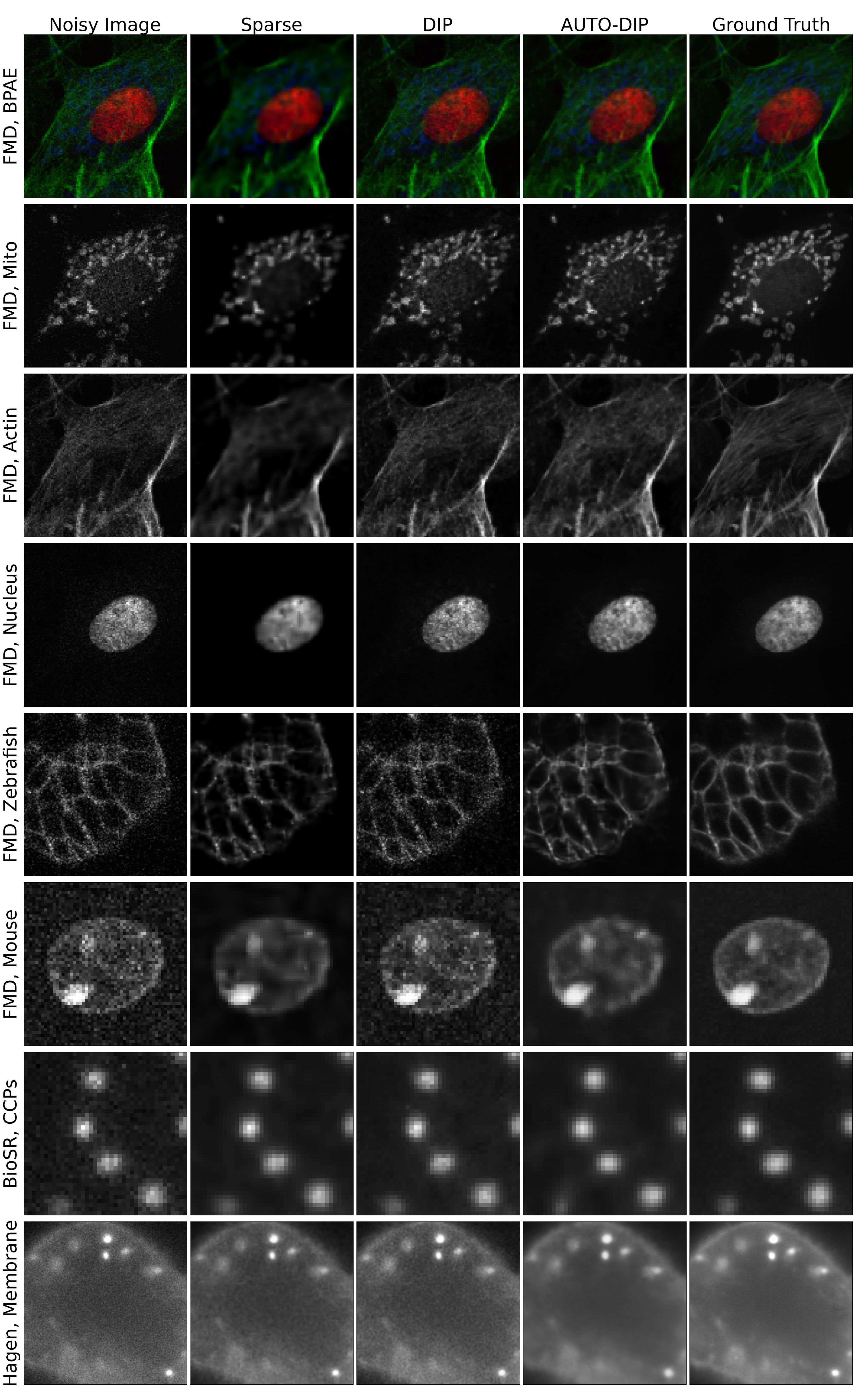}
		\caption{Generalization results for various test sets. From left to right: noisy input image, result from sparsity-based denoising, result with original DIP configuration, result with AUTO-DIP, ground truth image. First row: \ac{roi} of a confocal microscopy image of a fixed BPAE cell from the FMD dataset with three channels depicting mitochondria (red), actin (green), and nuclei (blue). Rows 2-4: single-channel grey-scale images of the above image. Rows 5-6: confocal microscopy of zebrafish embryo and two-photon microscopy of mouse brain tissue from the FMD dataset. Rows 7-8: widefield microscopy of CCPs and membrane from the BioSR, and Hagen datasets.\\
        Abbreviations: BPAE -- \acl{bpae}, CCPs -- \aclp{ccp}, Mito -- Mitochondria}
		\label{fig:test-results}
	\end{figure*}

\cref{fig:test-results} shows example images for the test sets of the public data sets. While denoising using the original DIP configuration retains a significant amount of noise, the results for the  proposed parameter transfer configuration are clearly superior for most cases. In particular images with dot-like structures become very similar to the ground truth (see the example from the BioSR dataset). However, fine line structures appear slightly blurred in some cases (e.g., the FMD example shown in the third row of \cref{fig:test-results}), and in some cases, an oversmoothing effect can be seen (sixth row). 
In comparison, the state-of-the-art sparsity-based variational approach also performs well for images with sparse foreground structures (e.g., fifth row), but struggles to denoise images with fine, dense structures (e.g., third row).  

The visual impressions are supported by the quantitative evaluation summarized in  \Cref{tab:results_overview}. The table contains the PSNR and LPIPS values for the publicly available test datasets, divided by sub-datasets. Overall, the transfer DIP configuration achieves the best performance, with consistent improvements in PSNR and SSIM over the original DIP and the noisy baseline. The largest gains relative to the original DIP occur for widefield acquisitions in the FMD and Hagen sets. The sparsity-based method is competitive for images with sparse structures well separated from the background, as in BioSR CCPs and FMD FISH, but otherwise performs worse than the DIP approaches.

\begin{figure*}
		\centering
		\includegraphics[width=\linewidth]{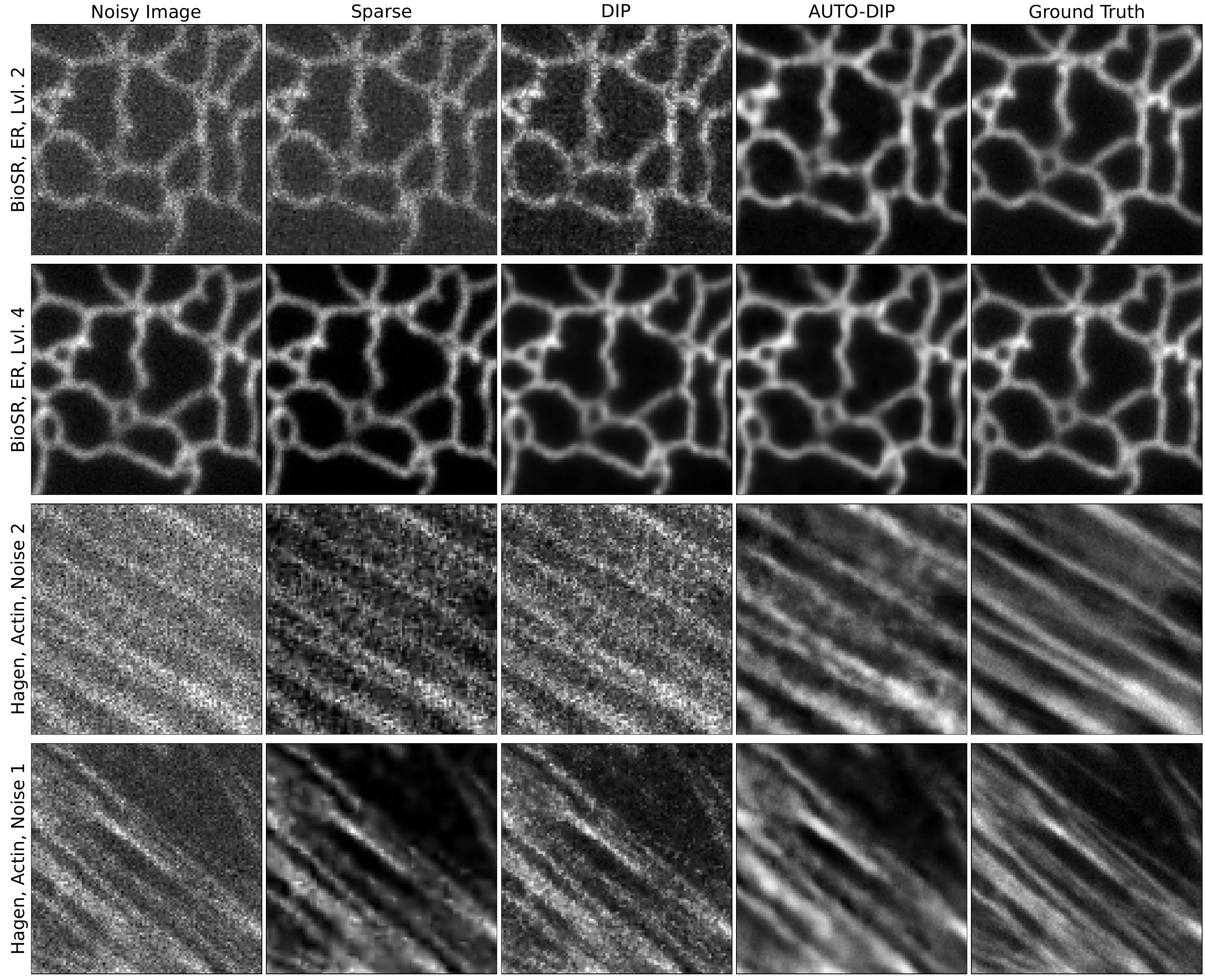}
		\caption{Impact of noise level on denoising performance illustrated for BioSR and Hagen data. From left to right: noisy input image; output of sparsity-based variational denoising; original DIP configuration; AUTO-DIP output; ground truth image. First two rows show images of endoplasmatic reticulum (ER) with high noise level (row 1) and low noise level (row 2). Last two rows show actin with high noise level (row 3) and low noise level (row 4).}
		\label{fig:noise_comp}
	\end{figure*}

Despite the overall denoising improvement, for a few cases only marginal differences between the original and transfer DIP exist, with the original DIP occasionally performing better (e.g., FMD Mito, Confocal). This can be explained by the noise conditions of the input data. As illustrated in \cref{fig:noise_comp}, which presents example images from the Hagen and BioSR datasets at low and high noise levels, all denoising approaches perform well for low noise images. In contrast, the transfer configuration produces high-quality results for both conditions, whereas for the high noise levels, substantial noise remains in outputs from the original DIP and the sparsity-based variational denoising approach.
 
\begin{figure*}
		\centering
		\includegraphics[width=0.95\linewidth]{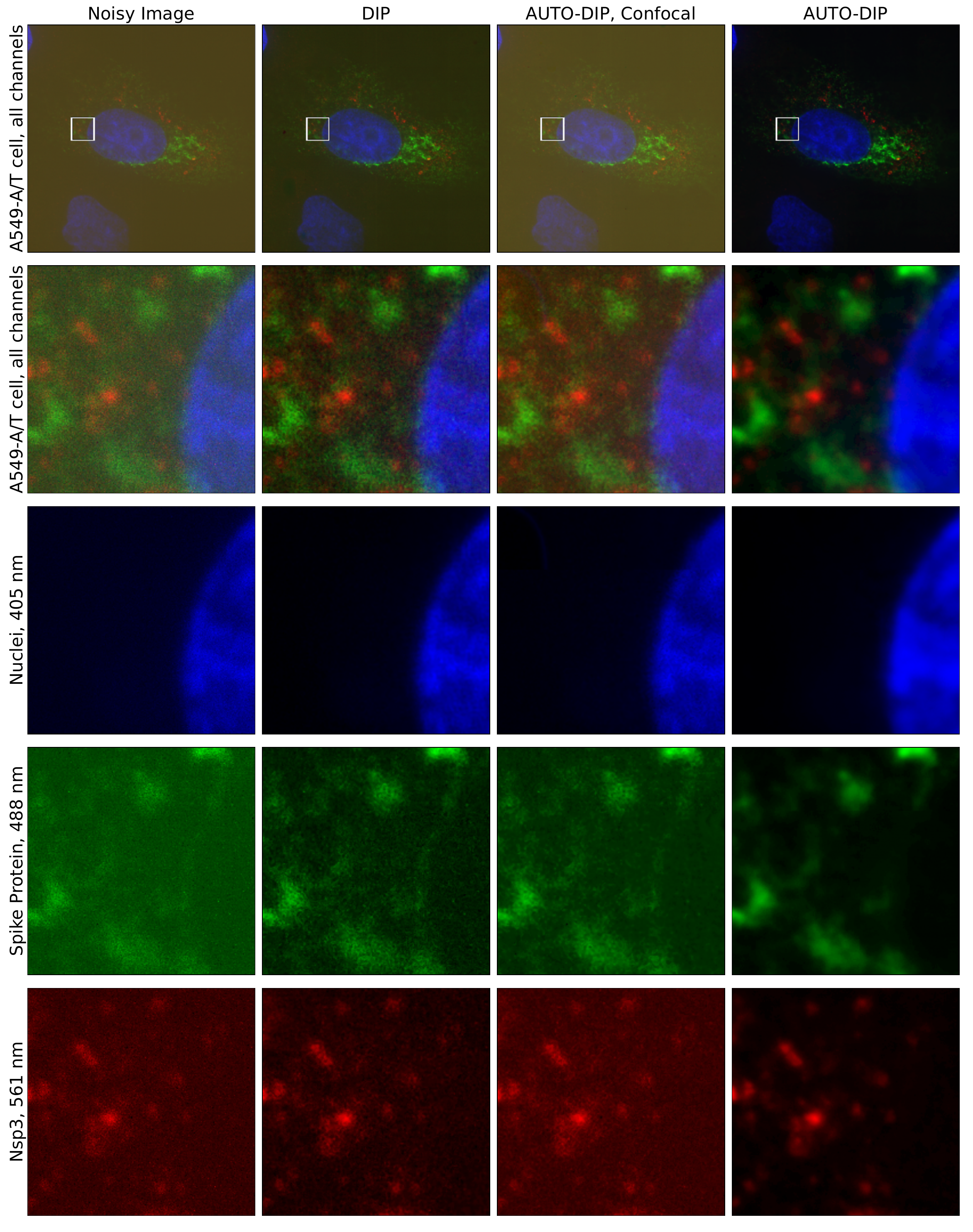}
		\caption{AUTO-DIP denoising results for an A549-A/T cell infected with SARS-CoV-2. From left to right: noisy input image; result with original DIP configuration; AUTO-DIP result with parameter transfer based on the microscope type (here: confocal microscope); AUTO-DIP result with parameter transfer based on UMAP-based semantic image similarity. First row: three channel image of the infected A549-A/T cell with marked \ac{roi}. Channels depict nuclei (blue, 405 nm), the Spike protein (green, 488 nm), and Nsp3 (red, 561 nm). Second row: \ac{roi} for the combined channels. Rows 3-5: single-channel images for the \ac{roi}.\\}
		\label{fig:covid}
	\end{figure*}

In addition to the public test sets, we also tested the proposed transfer strategy for the UKE internal data. For \cref{fig:covid}, two transfer variants were evaluated: the first one based on the microscope type and the second one based on the closest calibration image in the UMAP representation. Both the original configuration and the transfer configuration with the specified microscope type effectively reduce image noise. However, in this particular example, the transfer without specifying the microscope type achieves greater noise reduction but at the expense of some fine details.
		
	\begin{table*}
		\caption{Results for test sets subdivided by microscope type and specimen. For the FMD Test, BioSR, Shah, and W2S datasets, only results for the two highest noise levels are shown. From left to right: PSNR/LPIPS values for the noisy image, output from sparse denoising, output from the original DIP, and output from DIP with transferred parameters.}
		\renewcommand{\arraystretch}{1.2}
		\begin{tabularx}{\linewidth}{ p{1.3cm} p{4cm} | *{4}{>{\raggedleft\arraybackslash}X}}
			\toprule
			\raggedright\arraybackslash Dataset &
			\raggedright\arraybackslash Sub-Dataset &
			\raggedright\arraybackslash Noisy Image &
            \raggedright\arraybackslash Sparse Denoising &
			\raggedright\arraybackslash DIP &		
			\raggedright\arraybackslash AUTO-DIP\\ 
			\midrule
FMD Test & avg2 & 30.08 / 0.263 & 28.22 / 0.169 & 32.07 / 0.167 & \textbf{34.13} / \textbf{0.091} \\
FMD Test & raw & 27.22 / 0.364 & 27.43 / 0.199 & 28.97 / 0.264 & \textbf{32.38} / \textbf{0.119} \\
FMD Test & BPAE, Actin (G), Confocal & 29.89 / 0.191 & 26.85 / 0.161 & 31.70 / 0.106 & \textbf{31.90} / \textbf{0.094} \\
FMD Test & BPAE, Actin (G), Two-Photon & 26.34 / 0.380 & 28.38 / 0.228 & 27.60 / 0.257 & \textbf{29.83} / \textbf{0.183} \\
FMD Test & BPAE, Actin (G), Widefield & 25.42 / 0.383 & 27.51 / 0.305 & 25.98 / 0.359 & \textbf{30.98} / \textbf{0.184} \\
FMD Test & BPAE, Mito (R), Confocal & 34.53 / 0.071 & 32.29 / 0.085 & \textbf{36.57} / \textbf{0.031} & 36.50 / 0.034 \\
FMD Test & BPAE, Mito (R), Two-Photon & 32.34 / 0.239 & 29.09 / 0.179 & 35.57 / 0.091 & \textbf{36.52} / \textbf{0.074} \\
FMD Test & BPAE, Mito (R), Widefield & 27.77 / 0.469 & 27.09 / 0.388 & 28.51 / 0.400 & \textbf{33.81} / \textbf{0.199} \\
FMD Test & BPAE, Nucleus (B), Confocal & 33.41 / 0.136 & 31.30 / 0.047 & 36.08 / 0.028 & \textbf{37.04} / \textbf{0.015} \\
FMD Test & BPAE, Nucleus (B), Two-Photon & 25.55 / 0.386 & 25.34 / 0.135 & 27.69 / 0.242 & \textbf{29.92} / \textbf{0.063} \\
FMD Test & BPAE, Nucleus (B), Widefield & 26.94 / 0.470 & 25.86 / 0.374 & 28.18 / 0.393 & \textbf{34.38} / \textbf{0.126} \\
FMD Test & Fish, Confocal & 24.38 / 0.389 & 26.99 / \textbf{0.099} & 25.41 / 0.311 & \textbf{29.40} / 0.163 \\
FMD Test & Mice, Confocal & 30.90 / 0.250 & 27.84 / 0.070 & 34.48 / 0.080 & \textbf{36.59} / \textbf{0.029} \\
FMD Test & Mice, Two-Photon & 26.30 / 0.405 & 25.33 / 0.134 & 28.47 / 0.291 & \textbf{32.14} / \textbf{0.090} \\
Hagen & Actin, Confocal & 25.36 / 0.115 & 21.42 / 0.230 & 26.57 / \textbf{0.076} & \textbf{27.23} / 0.097 \\
Hagen & Actin, Widefield, 20x, noise1 & 24.44 / 0.501 & 28.39 / 0.261 & 26.51 / 0.392 & \textbf{30.30} / \textbf{0.099} \\
Hagen & Actin, Widefield, 60x, noise1 & 28.24 / 0.350 & 35.30 / 0.105 & 31.81 / 0.198 & \textbf{35.34} / \textbf{0.078} \\
Hagen & Actin, Widefield, 60x, noise2 & 18.49 / 0.606 & 19.36 / 0.579 & 18.82 / 0.660 & \textbf{23.11} / \textbf{0.521} \\
Hagen & Membrane, Widefield & 29.54 / 0.311 & 33.81 / 0.096 & 33.15 / 0.135 & \textbf{35.02} / \textbf{0.071} \\
Hagen & Mito, Confocal & 22.52 / 0.180 & \textbf{23.90} / 0.226 & 23.15 / 0.154 & 23.53 / \textbf{0.142} \\
Hagen & Mito, Widefield, 20x, noise1 & 24.97 / 0.474 & 27.56 / 0.271 & 26.63 / 0.415 & \textbf{30.11} / \textbf{0.084} \\
Hagen & Mito, Widefield, 60x, noise1 & 28.47 / 0.376 & 34.36 / 0.098 & 31.75 / 0.227 & \textbf{35.61} / \textbf{0.044} \\
Hagen & Mito, Widefield, 60x, noise2 & 20.14 / 0.490 & 21.20 / 0.550 & 20.47 / 0.553 & \textbf{24.32} / \textbf{0.401} \\
Hagen & Nucleus, Widefield & 24.81 / 0.492 & \textbf{34.14} / 0.173 & 28.71 / 0.258 & 33.11 / \textbf{0.168} \\
BioSR & CCPs, Widefield & 25.79 / 0.383 & \textbf{32.26} / \textbf{0.089} & 29.97 / 0.141 & 27.44 / 0.178 \\
BioSR & ER, Widefield & 18.40 / 0.626 & 21.06 / 0.450 & 19.57 / 0.527 & \textbf{22.65} / \textbf{0.265} \\
BioSR & F-actin, Widefield & 18.79 / 0.578 & 20.86 / \textbf{0.337} & 20.86 / 0.443 & \textbf{23.84} / 0.340 \\
BioSR & Microtubules, Widefield & 22.05 / 0.472 & 27.36 / 0.281 & 25.04 / 0.281 & \textbf{28.08} / \textbf{0.204} \\
Shah & Tubulin, Widefield & 28.10 / 0.171 & 23.57 / 0.154 & \textbf{30.13} / \textbf{0.144} & 29.18 / 0.353 \\
W2S & Human Cells, Widefield, Channel 1 & 29.28 / 0.432 & 30.39 / 0.165 & 32.94 / 0.276 & \textbf{37.07} / \textbf{0.040} \\
W2S & Human Cells, Widefield, Channel 2 & 29.21 / 0.438 & 31.71 / \textbf{0.125} & 32.53 / 0.285 & \textbf{35.38} / 0.153 \\
W2S & Human Cells, Widefield, Channel 3 & 29.44 / 0.508 & 30.70 / 0.246 & 32.91 / 0.348 & \textbf{38.11} / \textbf{0.069} \\

			\bottomrule
		\end{tabularx}
		\label{tab:results_overview}
	\end{table*}

	\section{Discussion}
 
\noindent The results demonstrate that AUTO-DIP makes DIP–based denoising feasible in fluorescence microscopy without the need for lengthy and computationally costly parameter tuning. The proposed approach is built on transferring architecture and stopping-point parameters from similar images of a dedicated calibration image set. Among the evaluated transfer methods, the most effective strategy was to assign each image the configuration that achieved the best average performance across all calibration images sharing the same microscope and specimen type. Using this transfer scheme, AUTO-DIP provides a strong initialization that reduces run time to roughly one-third of the time required by the standard DIP configuration. Compared to the denoising performance of the original DIP configuration, AUTO-DIP mostly achieved superior, often near-optimal results. This substantially broadens the practical usability of DIP in microscopy, especially when supervised learning is not applicable due to the lack of ground truth data. 
 

 
 Our analyses reveal several insights into DIP behavior in microscopy data. First, optimal architectures varied considerably, even between patches from the same image, underlining that universal architectures are not necessarily suitable for microscopy data. We observed a systematic trend: images with higher mean gradients tended to benefit from shallower but wider networks (up to width 128), while lower-gradient images performed best with deeper and narrower configurations. These findings align with those of Liu et al.~\cite{Liu_Li_Pang_Nie_Yap_2023} for natural images, who reported that fine-grained structures benefit from wide, shallow networks, while coarse-grained content favors narrow, deeper architectures. 
 Second, the required number of iterations was found to depend strongly on the microscopy modality: optimization for widefield microscopy images converged faster than for confocal or two-photon microscopy images. This might be explained by the higher noise levels in widefield microscopy imaging, which reduce the amount of recoverable detail and cause the network to overfit earlier. Third, AUTO-DIP proved especially effective for high noise levels, where the baseline DIP configuration and the sparsity-based variational denoising struggled. 

 When comparing our results to the literature, AUTO-DIP applied to the two highest noise levels of the FMD test set achieved PSNR values in the mid-to-upper range of those reported for classical denoising methods evaluated by Zhang et al.~\cite{Zhang_Zhu_Nichols_Wang_Zhang_Smith_Howard_2019}. However, supervised deep learning-based approaches reported by Zhang et al.~\cite{Zhang_Zhu_Nichols_Wang_Zhang_Smith_Howard_2019} and Lohr et al.~\cite{Lohr_Meyer_Woelk_Kovacevic_Diercks_Werner_2025} still outperform AUTO-DIP by approximately 2-\SI{3}{\decibel} in terms of PSNR. For the Hagen dataset, supervised deep learning methods evaluated by Hagen et al.~\cite{Hagen_Bendesky_Machado_Nguyen_Kumar_Ventura_2021} and Lohr et al.~\cite{Lohr_Meyer_Woelk_Kovacevic_Diercks_Werner_2025} achieve up to \SI{5}{\decibel} higher PSNR. For the W2S, BioSR, and Shah datasets, no benchmark values are currently available.
 These differences highlight that, while AUTO-DIP significantly improves the practicality of DIP for microscopy denoising, it still lags behind fully supervised approaches in terms of image restoration quality when sufficient data is available for supervised training of deep learning networks.

 
 Despite the promising results, the present study has some limitations, which will be adressed in future work to extend the capabilities of AUTO-DIP. For instance, the calibration dataset was relatively small, which restricted the diversity of parameter-transfer examples. Expanding the calibration dataset could improve parameter transfer by providing a wider variety of reference images, making AUTO-DIP less sensitive to the specific images selected. Further, a promising optimization direction could be to replace the conventional mean squared error (MSE) loss used during DIP optimization with perceptual or structural similarity measures, such as LPIPS, PSNR, or combinations thereof, which we already used for parameter selection in this study. These measures may better capture relevant image features than pixel-wise errors alone. Additional architectural modifications 
 may also reduce the blurring observed in some of the current results.  Furthermore, exploring alternative similarity measures will be important, as defining measures that accurately reflect perceptual similarity remains a challenging problem in image analysis.

	\FloatBarrier
	\bibliography{literature}

		
		
		
		
		
		
		
		
		
		

\end{document}